\documentclass{article} 
\usepackage{iclr2027_conference,times}

\usepackage{amsmath,amsfonts,bm}

\def\eqref#1{equation~\ref{#1}}

\def\1{\bm{1}}

\DeclareMathAlphabet{\mathsfit}{\encodingdefault}{\sfdefault}{m}{sl}
\SetMathAlphabet{\mathsfit}{bold}{\encodingdefault}{\sfdefault}{bx}{n}

\usepackage{hyperref}
\usepackage{url}
\usepackage{graphicx}
\usepackage{booktabs,multirow,graphicx,amssymb}
\usepackage[table]{xcolor}
\usepackage{wrapfig}
\usepackage{booktabs}
\usepackage{wrapfig}
\usepackage{tabularx}
\usepackage{graphicx}
\definecolor{oursblue}{RGB}{234,242,250}

\title{Spike-driven Vision–Language–Action Model}

\author{Shuai Wang$^{1,2}$, {Malu Zhang}$^{1,2}$\thanks{Corresponding author: maluzhang@uestc.edu.cn}, Mingquan Liu$^{1,3}$,\textbf{Weihui Dai}$^{1}$, Dehao Zhang$^{1}$,\\ \textbf{Jieyuan Zhang}$^{1}$, \textbf{Yimeng Shan}$^{1}$, \textbf{Zijian Zhou}$^{1}$, 
\textbf{Peng Zhou}$^{3}$, \textbf{Bolun Wang}$^{3}$, 
\textbf{Yang Yang}$^{1}$.
~\\
$^{1}$University of Electronic Science and Technology of China, \\
$^{2}$Shenzhen Loop Area Institute, 
$^{3}$LuxiTech Co. Ltd., Shenzhen, China.
}

\iclrfinalcopy

\begin{document}

\maketitle

\begin{abstract}
Vision–language–action (VLA) models bridge multimodal understanding and robotic control, advancing the dominant paradigm for embodied intelligence. However, most existing models rely on large Transformers, whose latency and energy costs hinder deployment on resource-constrained platforms. Through sparse event-driven computation, spiking neural networks offer a promising paradigm for high-performance and energy-efficient computing. Here, we propose the first Spike-driven VLA framework enabling end-to-end direct training for robotic manipulation, which mainly comprises three core components. First, we develop spiking visual and instruction encoders for multimodal perception, encoding visual observations and language instructions into sparse, reliable spike representations for subsequent cross-modal fusion. Then, we introduce Multi-Winner Spike Fusion for instruction-guided scene understanding, using bidirectional top-$k$ winner-take-all spike routing to suppress background interference and yield fused memory. 
Finally, we propose a Spike Action Chunking Transformer that incorporates spiking cross-attention over the fused memory and the current robot state, enabling efficient end-to-end generation of continuous action chunks for robotic control. Extensive experiments on LIBERO and Meta-World demonstrate that Spike-driven VLA achieves competitive performance with fewer parameters and lower estimated inference energy than conventional VLA models.  This work establishes a foundational framework for neuromorphic VLA modeling. 
\end{abstract}

\section{Introduction}

Vision–language–action (VLA) models~\citep{brohan2023rt2,kim2024openvla} enable robots to perceive their surroundings, interpret natural-language instructions, and generate corresponding actions within a unified policy, providing a promising foundation for general-purpose robotic control~\citep{black2024pi0}. However, many existing VLAs build on large pretrained vision–language backbones for instruction understanding and multimodal reasoning~\citep{brohan2023rt2,kim2024openvla,black2024pi0}. Their large model scale and dense computation impose substantial memory and latency overheads, hindering deployment on resource-constrained robotic platforms~\citep{wen2024tinyvla,shukor2025smolvla,budzianowski2025edgevla}. To this end, lightweight VLAs~\citep{wen2024tinyvla, xie2026turbovla,shukor2025smolvla} improve efficiency through compact architectures. However, they still rely on dense multiply–accumulate (MAC)~\citep{you2020shiftaddnet} operations, making repeated policy inference costly during closed-loop control. Therefore, developing compact and efficient VLAs for edge deployment remains an important research direction.

Spiking Neural Networks (SNNs)~\citep{maass1997networks,gerstner2002spiking} offer a promising alternative for energy-efficient intelligence through brain-inspired dynamics and spike-driven computation~\citep{chen2023hybrid}.
Unlike Artificial Neural Networks (ANNs)~\citep{he2016deep}, which primarily rely on dense MAC operations~\citep{you2020shiftaddnet}, spiking neurons~\cite{izhikevich2003simple, wangbipolar, wang2025ternary} trigger synaptic computation only when spikes arrive, replacing multiplications with sparse accumulate (AC) operations~\citep{bouvier2019spiking,roy2019towards}. An AC operation typically consumes only 1/5 to 1/10 of the energy required by a MAC operation on a representative 45-nm ASIC~\citep{you2020shiftaddnet}, while dedicated hardware platforms can further exploit sparse spike activity~\citep{deng2020tianjic}. These properties make SNNs a promising computational foundation for efficient VLA deployment on resource-constrained edge platforms.

\begin{wrapfigure}{r}{0.55\textwidth}
    \centering
    \includegraphics[width=0.55\textwidth]{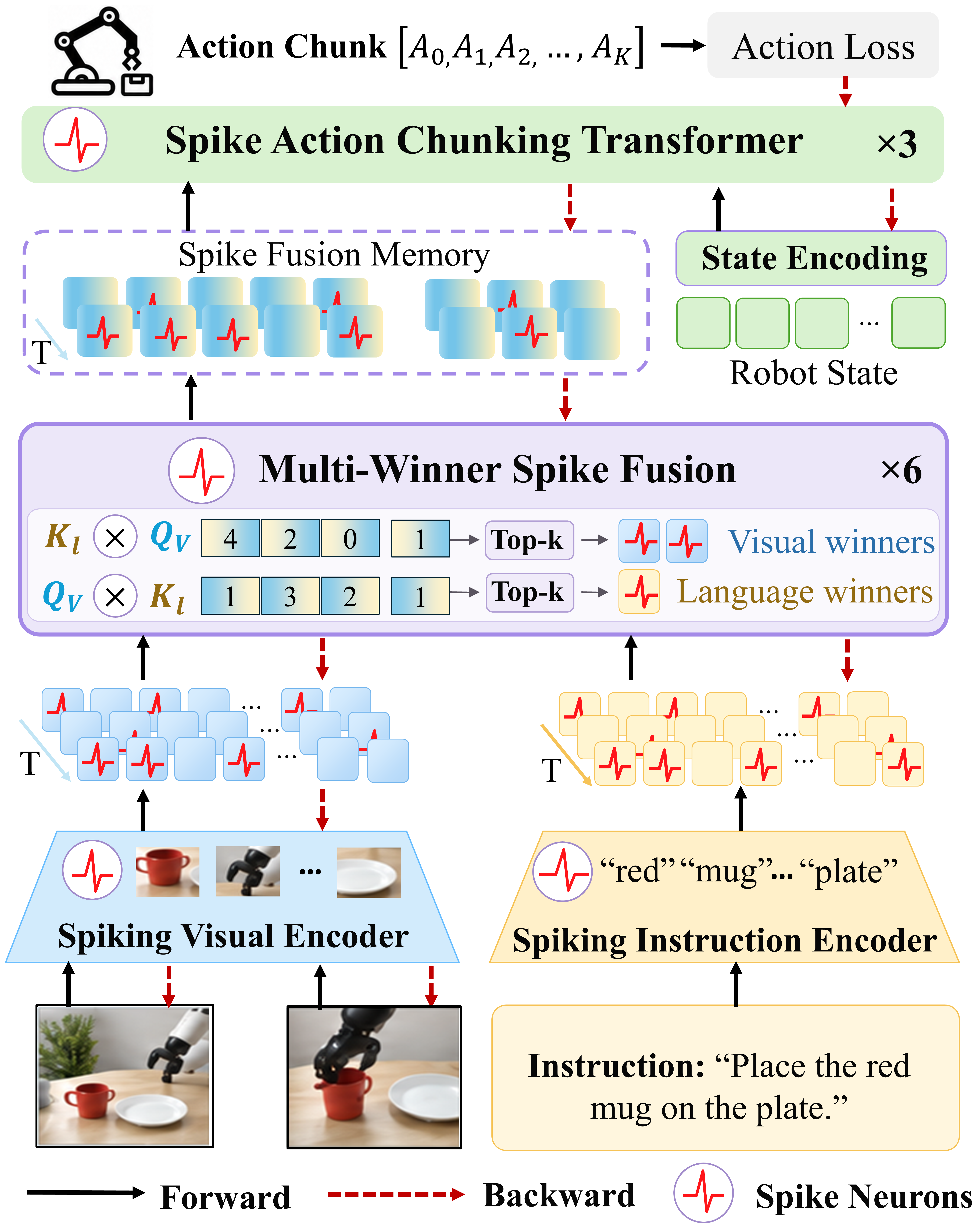}  
    \caption{Overview of Spike-driven VLA.}
    \label{Fig1}
    \vspace{-10pt}
\end{wrapfigure}
Existing spiking VLAs~\cite{song2026spikevla} primarily rely on ANN-to-SNN conversion~\cite{li2021free, you2024spikezip}, transferring the representations learned by pretrained ANNs to their spiking counterparts. However, ANN-to-SNN methods~\citep{chen2026loss} typically retain the source ANN architecture, restricting task-specific architectural design. They also approximate continuous activations by accumulating spikes~\citep{wang2026training} over multiple timesteps, resulting in substantial temporal overhead~\cite{wang2026training}. In contrast, direct training methods~\cite{wu2018spatio,zhou2023spikformer} enable task-specific spiking architectures and end-to-end learning of task-relevant representations in the spike domain. Recent advances in directly trained SNNs span perception tasks, including classification~\citep{cai2024eeg}, detection~\citep{yao2025scaling}, and tracking~\citep{wang2024pssd,shan2025sdtrack}, as well as language modeling~\citep{zhu2023spikegpt} and robotic control~\citep{zhang2025toward}. Despite these advances, extending directly trained SNNs to end-to-end VLA modeling for robotic manipulation remains challenging.

To break the ice, we propose the first Spike-driven VLA framework that supports end-to-end direct training for robotic manipulation. As shown in Fig.~\ref{Fig1}, Spike-driven VLA comprises three stages: multi-modal spike encoding, cross-modal fusion, and state-conditioned action decoding. First, we propose a spiking vision encoder (SVE) and a spiking instruction encoder (SIE) to capture visual context and instruction semantics in modality-specific spike representations. Then, we propose a Multi-Winner Spike Fusion (MWSF) module that establishes task-relevant cross-modal correspondence, yielding action-ready fusion memory. Finally, based on this memory and the current robot state, we propose Spike Action Chunking Transformer (SpikeACT) to generate continuous action chunks for instruction-guided manipulation.
We evaluate Spike-driven VLA on the LIBERO~\citep{liu2023libero} and Meta-World~\citep{yu2020meta} benchmarks, which span diverse language-conditioned manipulation tasks and simulated environments. Extensive experiments demonstrate that Spike-driven VLA achieves competitive performance against representative ANN-based methods while requiring fewer parameters and computational operations. Further evaluation on LIBERO-Plus~\citep{fei2026liberoplus} demonstrates its strong robustness under diverse distribution shifts. These results demonstrate that our Spike-driven VLA offers a new solution for efficient VLA deployment on resource-constrained edge platforms. The main contributions are as follows:

\begin{itemize}
    \item We develop an SVE and a BERT-based SIE for multimodal perception. These encoders capture hierarchical visual features and contextual instruction semantics, respectively, and preserve them in sparse spike representations.

    \item  We propose MWSF to fuse visual and instruction features into a memory for scene understanding. Its bidirectional top-$k$ winner-take-all spike routing suppresses background interference and reduces cross-modal computation through sparse token aggregation.

    \item  We develop the SpikeACT module for multimodal decision-making and continuous control. It combines fused memory with the current robot state to jointly predict action chunks, enabling efficient action generation through sparse spike-driven computation.

    \item  Building on these components, we propose a Spike-driven VLA framework that supports end-to-end direct training for robotic manipulation. Experiments on LIBERO and Meta-World demonstrate competitive performance with fewer parameters and operations, offering a new path toward efficient VLA deployment on resource-constrained edge platforms.
\end{itemize}

\section{Related Work}
\subsection{Spiking Neural Networks}

Benefiting from sparse event-driven computation, SNNs offer a promising paradigm for energy-efficient computing on resource-constrained edge devices. However, early SNNs suffer from limited performance due to immature training methods and shallow architectures. Advances in spatio-temporal backpropagation~\citep{fang2021incorporating} have substantially facilitated the training of high-performance SNNs. Building on these advances, Spikformer~\citep{zhou2023spikformer} established the spiking self-attention paradigm, while SDT-V1/V2~\citep{yao2023spikedriven,yao2024spike} reformulated self-attention with the Hadamard product. Subsequent studies have extended Spiking Transformers to diverse vision tasks, such as classification~\citep{zheng2021high, ICLR2025_3c85f70d,ICLR2025_3c85f70d}, detection~\citep{yao2025scaling}, and tracking~\citep{shan2025sdtrack}, demonstrating their potential as general spiking backbones. Beyond vision, SNNs have also made substantial progress in language modeling. SpikeGPT~\citep{zhu2023spikegpt} demonstrated generative language modeling with directly trained SNNs. SpikeBERT~\citep {liangspikinglm, zhousmoothspike} advanced language understanding through knowledge distillation and direct training, respectively. Collectively, advances in spiking perception and language modeling, together with recent progress in spiking robotic control~\citep{zhang2025toward}, enable the end-to-end direct training of a fully Spike-driven VLA framework.

\subsection{Vision–Language–Action Models}
VLA models have emerged as an important paradigm for embodied intelligence. RT-2~\citep{brohan2023rt2} transfers knowledge acquired through vision-language pretraining to robotic control, while PaLM-E~\citep{driess2023palme} extends multimodal language modeling to embodied decision-making. Built upon the Open X-Embodiment dataset, RT-X~\citep{o2024open} further advances cross-robot policy learning. More recent models, including Octo~\citep{octo2024}, OpenVLA~\citep{kim2024openvla}, and $\pi_0$~\citep{black2024pi0}, improve task generalization by scaling robot data and multimodal pretraining. Despite their strong performance, these models rely heavily on large pretrained backbones and computationally intensive action-generation pipelines, imposing substantial memory, computational, and latency overhead for deployment on resource-constrained robotic platforms.  
Existing efforts improve VLA efficiency through action chunking and lightweight model design. Action chunking predicts multiple future actions in a single inference pass, spreading the computational cost across successive control steps. Meanwhile, TinyVLA~\citep{wen2024tinyvla}, TurboVLA~\citep{xie2026turbovla}, and SmolVLA~\citep{shukor2025smolvla} reduce inference costs through compact architectures and efficient backbones. However, these methods improve efficiency primarily through architectural compression, while leaving the underlying dense computation paradigm unchanged. Therefore, how to make full use of the event-driven nature of SNNs offers a promising path toward efficient VLA models.  

\section{Preliminaries}
\subsection{Spiking Neuron Model}
Spiking neurons replicate the behavior of biological neurons that integrate input spikes into their membrane potential $u$ and fire a spike only when $u$ exceeds a threshold. For layer \(l\) at timestep \(t\), the synaptic input is given by
\(I^l[t]=W^l s^{l-1}[t]\), where \(W^l\) denotes the synaptic weight matrix and \(s^{l-1}[t]\in\{0,1\}\) represents the binary spike activity from the preceding layer. Since only active presynaptic neurons contribute to \(I^l[t]\), synaptic computation can be implemented through sparse accumulation of the corresponding weights. We adopt the discrete Leaky Integrate-and-Fire (LIF) model~\citep{maass1997networks,gerstner2002spiking} denoted by $(\mathcal{SN}(\cdot))$, formulated as:
\begin{align}
\tilde{u}^l[t] = \lambda u^l[t-1] + I^l[t], \quad
s^l[t] = H\!\left(\tilde{u}^l[t]-V_{\mathrm{th}}\right), \quad
u^l[t] = \tilde{u}^l[t]-V_{\mathrm{th}}s^l[t],
\end{align}
where $\tilde{u}^l[t]$ and $u^l[t]$ denote the pre- and post-reset membrane potentials, respectively; $H(\cdot)$ is the Heaviside step function; $V_{\mathrm{th}}$ is the firing threshold; and $\lambda\in[0,1]$ controls membrane leakage. Since $H(\cdot)$ is non-differentiable at the threshold and has zero derivatives elsewhere, we approximate its derivative with a surrogate gradient during backpropagation to enable end-to-end training. Common choices~\cite{wu2018spatio} include derivatives of sigmoid, fast sigmoid, and arctangent functions. 

\subsection{Spike Self-attention}
Building on Vanilla Self-attention (VSA)~\cite{c:22} in ANNs, Spikformer~\citep{zhou2023spikformer} introduced Spiking Self-attention (SSA). For an input matrix $\mathbf{X}\in\mathbb{R}^{n\times d}$, queries $\mathbf{Q}$, keys $\mathbf{K}$, and values $\mathbf{V}$ are first computed via learnable weight matrices and then converted into spike trains by binary spiking neurons for subsequent processing. The $\mathbf{Q}$, $\mathbf{K}$ and $\mathbf{V}$ in SSA can be described as:
\begin{equation}
\mathbf{Q} = \mathcal{SN}(\text{BN}(\mathrm{Linear}(\mathbf{X}))),\quad  \mathbf{K} = \mathcal{SN}(\text{BN}(\mathrm{Linear}(\mathbf{X}))), \quad \mathbf{V} = \mathcal{SN}(\text{BN}(\mathrm{Linear}(\mathbf{X}))),
\end{equation}
where $\mathbf{Q}, \mathbf{K}, \mathbf{V} \in \mathbb{R}^{T \times n \times d}$, $\text{BN}(\cdot)$ denotes batch normalization and $\mathrm{Linear}(\cdot)$ refers to a linear operation. Unlike VSA, SSA omits the Softmax operation while retaining scaling to control the magnitude of the attention output. The attention output $\mathbf{Attn}$ is obtained by performing a matrix multiplication of the spiking $\mathbf{Q}$ and $\mathbf{K}$, scaled by factor \( s \), and then converted into spike trains:
\begin{equation}
\label{eq:SSA attn}
    \mathbf{Score} = s \cdot \mathbf{Q} \times \mathbf{K^{T}}, \quad
     \mathbf{Attn} = \mathcal{SN}(\mathbf{Score} \times \mathbf{V}).
\end{equation}
Therefore, SSA provides an energy-efficient self-attention computation paradigm without softmax normalization. It allows the ordering of the Q, K, and V matrices to be flexibly adjusted as needed, enabling the Spiking Transformer to capture global dependencies efficiently.

\section{Spike-driven Vision–Language–Action Models}
\subsection{Overall Architecture}
Spike-driven VLA consists of three functional stages:
(i) spiking visual and instruction encoders (SVE and SIE) for environment perception, detailed in Section~\ref{SVE};
(ii) a Multi-Winner Spike Fusion module for instruction-guided visual-language interaction, detailed in Section~\ref{MWSF};
and (iii) a SpikeACT module for chunked
control generation, detailed in Section~\ref{SpikeACT}. At control step $n$, the current visual observation $\mathcal O_n$ and language instruction $\mathcal L$ are encoded and integrated
through the MWSF module:
\begin{equation}
\begin{aligned}
\mathbf Z_{v,n}=E_v(\mathcal O_n),\quad
\mathbf Z_\ell=E_\ell(\operatorname{Tok}(\mathcal L)),\quad
\mathcal M_n=\mathcal F(\mathbf Z_{v,n},\mathbf Z_\ell).
\end{aligned}
\label{eq:spike_vla_perception_fusion}
\end{equation}
Here, $E_v$ and $E_\ell$ denote SVE and SIE, respectively.
They produce spike representations
$\mathbf Z_{v,n}\in\mathbb R^{N_v\times d}$ and
$\mathbf Z_\ell\in\mathbb R^{N_\ell\times d}$,
where $N_v$ and $N_\ell$ are the visual and instruction token counts,
and $d$ is their shared feature dimension.
$\operatorname{Tok}(\mathcal L)$ contains $N_\ell$ token indices.
$\mathcal F$ performs competitive visual-language interaction and
produces the task-conditioned memory
$\mathcal M_n\in\mathbb R^{(N_v+N_\ell)\times d}$.
The memory retains continuous membrane-potential residuals. The robot state $\mathcal S_n\in\mathbb R^{d_s}$ is then encoded and jointly processed with the fused memory:
\begin{equation}
\begin{aligned}
\mathbf Z_{s,n}=E_s(\mathcal S_n), \quad
\widehat{\mathcal A}_n =\operatorname{SpikeACT}
\left(\mathbf P_a,[\mathcal M_n;\mathbf Z_{s,n}]\right)
\in\mathbb R^{K\times d_a}.
\end{aligned}
\label{eq:spike_vla_state_action}
\end{equation}
Specifically, the State Encoder \(E_s\) maps the current robot state into \(N_s\) state tokens \(\mathbf Z_{s,n}\in\mathbb R^{N_s\times d}\), which are concatenated with the fused memory. Conditioned on $[\mathcal M_n;\mathbf Z_{s,n}]$, SpikeACT updates \(K\) learnable action tokens \(\mathbf P_a\in\mathbb R^{K\times d}\) through spiking cross-attention.The \(K\) decoded action tokens are aggregated across timesteps and projected to produce a continuous action chunk, \(\widehat{\mathcal A}_n=[\widehat{\mathbf a}_n,\ldots,\widehat{\mathbf a}_{n+K-1}]\), where each \(\widehat{\mathbf a}_{n+k}\in\mathbb R^{d_a}\). The individual components are detailed below.

\begin{figure*}[!htpb]
  \centering
  \includegraphics[width=\linewidth]{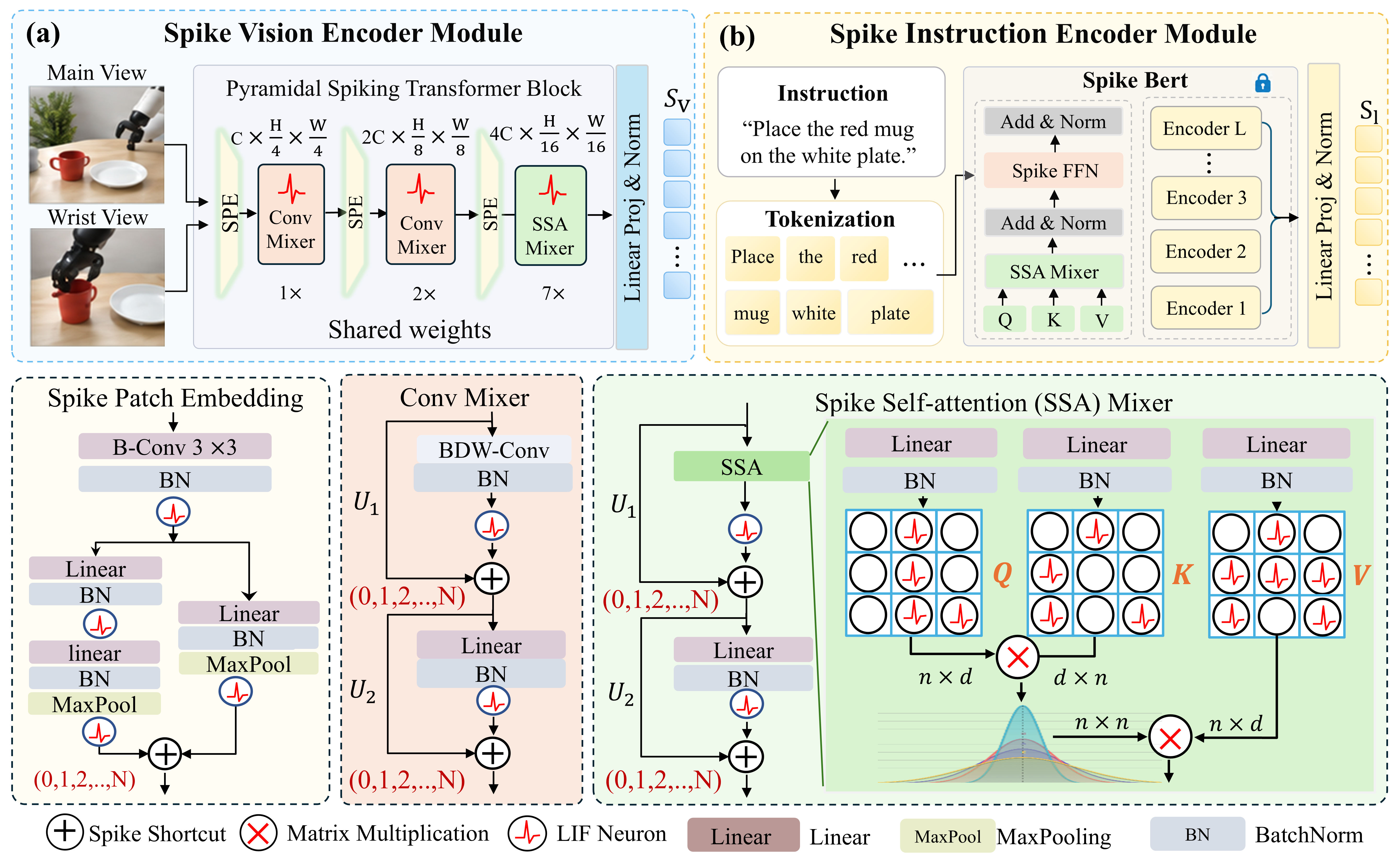}
\caption{
(a) SVE encodes main- and wrist-view observations into visual spike tokens capturing appearance and spatial relations; (b) SIE encodes language instructions into contextual semantic spike tokens that condition visual perception.
} 
\label{fig2}
\vspace{-5pt}
\end{figure*}

\subsection{Spike-driven Visual and Instruction Encoders}
\label{SVE}
The perception stage of Spike-driven VLA comprises SVE and SIE, which transform visual observations and language instructions into spike token representations. As shown in Fig.~\ref{fig2}(a), SVE encodes the main and wrist observations $\mathcal{O}_n^{\mathrm{Main}}$ and $\mathcal{O}_n^{\mathrm{wrist}}$, which can be described as:
\begin{equation}
\begin{aligned}
\mathbf{Z}_{v,n}
= E_v(\mathcal{O}_n^{\mathrm{Main}}, \mathcal{O}_n^{\mathrm{wrist}})
=\operatorname{BN}( \operatorname{Linear}_{v}\!\left(
\left[\mathbf{Z}^{\mathrm{main}};
\mathbf{Z}^{\mathrm{wrist}}\right]
\right)), \quad \mathbf{Z}_{v,n}\in\mathbb{R}^{T\times B\times N_v\times d}.
\end{aligned}
\label{eq:sve_encoding}
\end{equation}
where $\mathbf{Z}^{\mathrm{main}}$ and $\mathbf{Z}^{\mathrm{wrist}}$
denote dual-view features from a shared spiking visual transformer~\citep{yao2025scaling, zhou2024qkformer, wangbipolar}. The backbone is pretrained on ImageNet~\citep{deng2009imagenet} and subsequently fine-tuned during Spike-driven VLA training. The first two stages of the backbone contain one and two Conv Mixer blocks, respectively, which extract local features through convolution and downsampling. The final stage uses SSA modules. These updates enrich local visual features with global context, producing visual tokens that encode object appearance and spatial
relationships for subsequent instruction-guided fusion. Detailed architectural configurations are provided in Appendix~\ref{app:sve}. As shown in Fig.~\ref{fig2}(b), SIE encodes the instruction to provide semantic guidance for task-relevant visual perception, which can be described as:
\begin{equation}
\begin{aligned}
\mathbf{Z}_{\ell}
=E_{\ell}\left(\operatorname{Tok}(\mathcal{L})\right)
=\operatorname{BN}\left(
\operatorname{Linear}_{\ell}\left(
\mathbf{Z}^{\mathrm{text}}
\right)\right),
\quad
\mathbf{Z}_{\ell}\in\mathbb{R}^{T\times B\times N_{\ell}\times d}.
\end{aligned}
\label{eq}
\end{equation}
where $\mathbf{Z}^{\mathrm{text}}$ denotes contextual language features
extracted by a spiking BERT~\citep{liangspikinglm,zhousmoothspike} backbone. The backbone is pretrained with masked language modeling~\citep{zhousmoothspike} and remains frozen during Spike-driven VLA training, providing fixed semantic features for each instruction.
Detailed architectural configurations of Spiking BERT are provided in Appendix~\ref{app:sie}. SVE and SIE reduce intermediate activation memory while preserving control-relevant information, providing informative representations for subsequent scene understanding.

\begin{figure*}[!htpb]
  \centering
  \includegraphics[width=\linewidth]{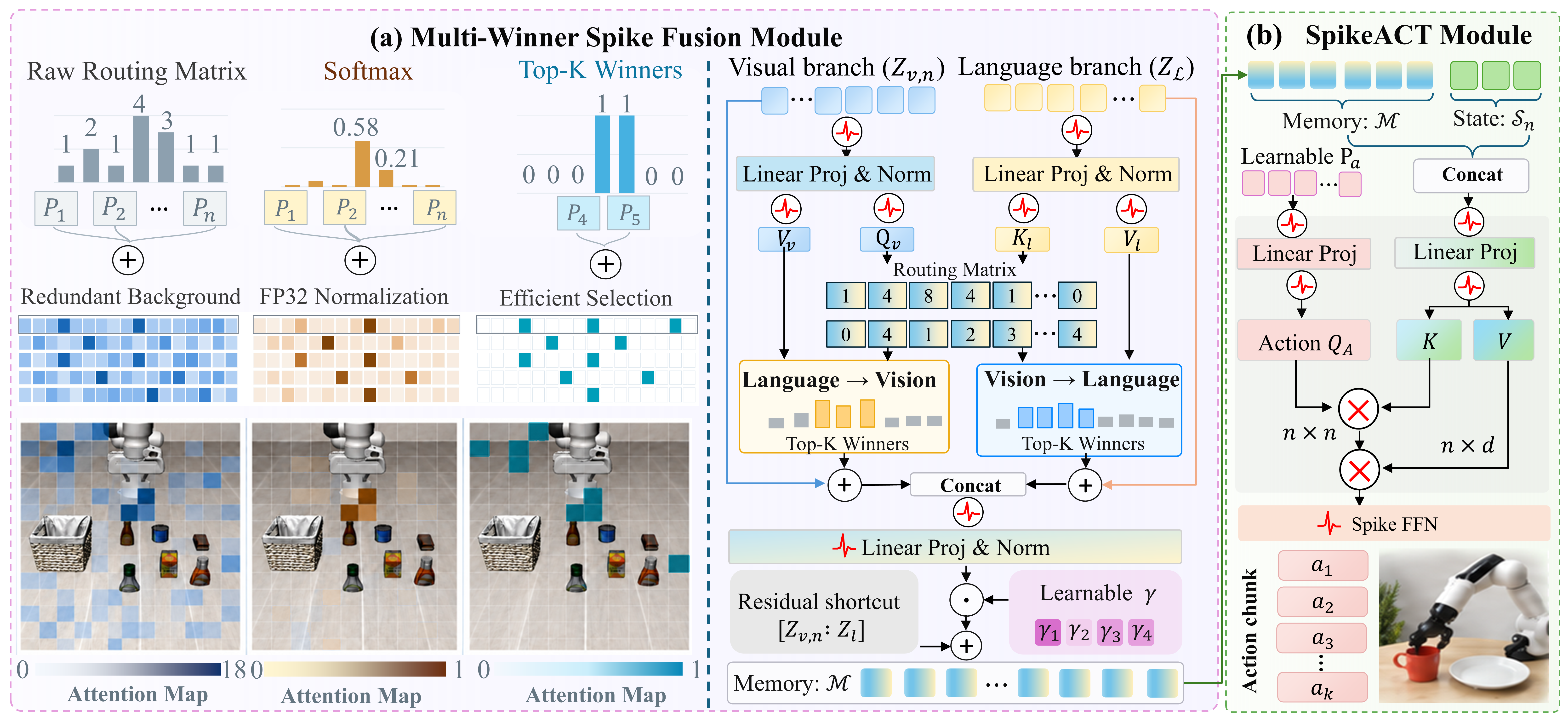}
\caption{
(a) MWSF uses bidirectional Top-$k$ WTA routing to retain instruction-relevant visual evidence, suppress background interference, and reduce computation. (b) SpikeACT uses learnable action tokens to predict continuous action chunks in parallel from fused memory and the robot state.
}
\vspace{-5pt}
  \label{fig:3}
\end{figure*}

\subsection{Multi-Winner Spike Fusion}
\label{MWSF}
Although SVE and SIE provide effective visual and language representations, precise robotic manipulation requires language instructions to direct visual perception toward task-relevant evidence. In robotic manipulation tasks, visual tokens typically far outnumber language tokens and contain substantial background clutter. Directly applying SSA~\citep{zhou2023spikformer} to global cross-modal interaction allows every visual token with a nonzero affinity to contribute to feature aggregation, causing weak background responses to accumulate and dilute salient evidence. To address this issue, we introduce the MWSF module, which combines a shared cross-modal affinity matrix with bidirectional competitive routing to establish sparse, task-relevant visual-language correspondences. As shown in Fig.~\ref{fig:3}(a), given $\mathbf{Z}_{v,n}$ and $\mathbf{Z}_{\ell}$, modality-specific spiking projections generate the features used for cross-modal matching and value aggregation:
\begin{equation}
\begin{aligned}
\mathbf{Q}_v &=
\mathcal{SN}\!\left(
\operatorname{BN}\!\left(
\operatorname{Linear}\!\left(
\mathcal{SN}(\mathbf{Z}_{v,n})
\right)\right)\right),
\quad
\mathbf{V}_v =
\mathcal{SN}\!\left(
\operatorname{BN}\!\left(
\operatorname{Linear}\!\left(
\mathcal{SN}(\mathbf{Z}_{v,n})
\right)\right)\right),\\
\mathbf{K}_{\ell} &=
\mathcal{SN}\!\left(
\operatorname{BN}\!\left(
\operatorname{Linear}\!\left(
\mathcal{SN}(\mathbf{Z}_{\ell})
\right)\right)\right),
\quad
\mathbf{V}_{\ell} =
\mathcal{SN}\!\left(
\operatorname{BN}\!\left(
\operatorname{Linear}\!\left(
\mathcal{SN}(\mathbf{Z}_{\ell})
\right)\right)\right).
\end{aligned}
\label{eq:fusion_projection}
\end{equation}
Here, $\mathbf{Q}_v,\mathbf{V}_v
\in\mathbb{S}^{T\times B\times N_v\times d}$ and
$\mathbf{K}_{\ell},\mathbf{V}_{\ell}
\in\mathbb{S}^{T\times B\times N_{\ell}\times d}$, $\mathbb{S}=\{0,1\}$ denotes the binary spike domain. Each $\operatorname{Linear}$
and $\operatorname{BN}$ is independently parameterized. The $\mathbf{Q}_v$ and $\mathbf{K}_{\ell}$ form a shared
cross-modal affinity matrix
$\mathbf{R}_n=\mathbf{Q}_v\mathbf{K}_{\ell}^{\top}
\in\mathbb{Z}^{T\times B\times N_v\times N_{\ell}}$.
The visual and language branches perform Top-$k$ multi-winner selection ($\operatorname{WTA}_{k}$) over
$\mathbf{R}_n$ and $\mathbf{R}_n^{\top}$. The $\operatorname{WTA}_{k}$ operator acts row-wise on $\mathbf R\in\mathbb Z^{T\times B\times N_q\times N_k}$,
where $N_q$ and $N_k$ are the numbers of $\mathbf{Q}_v$ and $\mathbf{K}_{\ell}$ tokens.
For any row $\mathbf r$ and its output $\mathbf a^{(k)}$, the $j$-th mask entry is defined as:
\begin{equation}
a_j^{(k)}=
\begin{cases}
1, & j\in\operatorname{TopK}_{k}(\mathbf r)
     \text{ and } r_j>0,\\
0, & \text{otherwise}.
\end{cases}
\label{eq:wta_selection}
\end{equation}
Here, $\operatorname{TopK}_{k}(\mathbf r)$ returns the indices of up to $k$ highest-scoring valid keys. As shown in Fig.\ref{fig:3}(a), WTA concentrates routing on instruction-relevant visual regions, suppressing background responses present in raw routing without requiring softmax normalization. The resulting cross-modal contexts are computed as follows:
\begin{equation}
\begin{aligned}
\mathbf{C}_v=
\mathbf{Z}_{v,n}+
\operatorname{BN}\!\left(
\operatorname{Linear}\!\left(
\mathcal{SN}\!\left(
\operatorname{WTA}_{k_v}(\mathbf{R}_n)\mathbf{V}_{\ell}
\right)\right)\right),\quad
\mathbf{C}_v \in\mathbb{R}^{T\times B\times N_v\times d}.
\end{aligned}
\label{eq:visual_context}
\end{equation}
\begin{equation}
\begin{aligned}
\mathbf{C}_{\ell}=
\mathbf{Z}_{\ell}+
\operatorname{BN}\!\left(
\operatorname{Linear}\!\left(
\mathcal{SN}\!\left(
\operatorname{WTA}_{k_{\ell}}(\mathbf{R}_n^{\top})\mathbf{V}_v
\right)\right)\right), \quad
\mathbf{C}_{\ell}
\in\mathbb{R}^{T\times B\times N_{\ell}\times d}.
\end{aligned}
\label{eq:bidirectional_wta}
\end{equation}
where \(\mathbf{C}_v\) denotes the instruction-guided visual representation, while \(\mathbf{C}_{\ell}\) denotes the visually grounded instruction representation. The two branches employ \(k_v\) and \(k_{\ell}\) as direction-specific routing budgets to accommodate the imbalance between visual and language sequence lengths. Finally, we concatenate $\mathbf{C}_v$ and $\mathbf{C}_{\ell}$ along the token
dimension to obtain the scene memory $\mathcal{M}_n$:
\begin{equation}
\begin{aligned}
\mathcal{M}_n
={}
[\mathbf{C}_v;\mathbf{C}_{\ell}]+
\boldsymbol{\gamma}\odot
\operatorname{BN}\!\left(
\operatorname{Linear}\!\left(
\mathcal{SN}\!\left(
[\mathbf{C}_v;\mathbf{C}_{\ell}]
\right)\right)\right), \quad \mathcal{M}_n\in
\mathbb{R}^{T\times B\times(N_v+N_{\ell})\times d}.
\end{aligned}
\label{eq:fused_memory}
\end{equation}
The residual connection preserves the cross-modal contexts obtained
through bidirectional routing and provides a direct pathway for
information and gradient propagation across stacked MWSF blocks,
preventing sparse routing updates from overwriting the original
representations. The learnable channel-wise scale
$\boldsymbol{\gamma}\in\mathbb{R}^{1\times d}$ adaptively controls the
magnitude of the residual update. This scaling does not compromise
spike-driven computation because the learnable projection operates on
the spike representation produced by $\mathcal{SN}$. Moreover,
$\boldsymbol{\gamma}$ can be folded into the affine scale and bias of
$\operatorname{BN}(\cdot)$ during inference, introducing no additional
non-spiking operation or runtime overhead. The resulting
$\mathcal{M}_n\in
\mathbb{R}^{T\times B\times(N_v+N_{\ell})\times d}$ remains continuous
and serves as the task-conditioned memory for subsequent
spiking action decoding.

\subsection{Spike Action Chunking Transformer}
\label{SpikeACT}
As described in the previous section, \(\mathcal{M}_n\) encodes scene context aligned with the current instruction. Mapping this memory to executable continuous control commands further requires conditioning on the robot’s current state. To this end, we develop SpikeACT module as shown in Fig.\ref{fig:3}(b), which employs both the scene memory and robot state and predicts a \(K\)-step action chunk in parallel. Specifically, the state encoder produces tokens $\mathbf{Z}_{s,n}\in\mathbb{R}^{T\times B\times N_s\times d}$.
These tokens are concatenated with $\mathcal{M}_n$ to form the
state-augmented memory $\widetilde{\mathcal{M}}_n$:
\begin{equation}
    \widetilde{\mathcal{M}}_n=[\mathcal{M}_n;\mathbf{Z}_{s,n}]_{\mathrm{token}},  \quad\widetilde{\mathcal{M}}_n\in
\mathbb{R}^{T\times B\times N_m\times d},
\end{equation}
where \(N_m=N_v+N_\ell+N_s\). We use $K$ learnable action 
$\mathbf{P}_a\in\mathbb{R}^{K\times d}$, each corresponding to one
future position in the action chunk. Independent spiking projections then generate the query features from
$\mathbf{P}_a$ and the key and value features from
$\widetilde{\mathcal{M}}_n$:
\begin{equation}
\begin{aligned}
\mathbf{Q}_a
=
\mathcal{SN}\!\left(
\operatorname{BN}\!\left(
\operatorname{Linear}\!\left(
\mathcal{SN}(\mathbf{P}_a)
\right)\right)\right),  \quad
\mathbf{K_{\widetilde{\mathcal{M}}_n},V_{\widetilde{\mathcal{M}}_n}}
=
\mathcal{SN}\!\left(
\operatorname{BN}\!\left(
\operatorname{Linear}\!\left(
\mathcal{SN}(\widetilde{\mathcal{M}}_n)
\right)\right)\right),
\end{aligned}
\label{eq:action_qkv}
\end{equation}
where
$\mathbf{Q}_a\in\mathbb{R}^{T\times B\times K\times d}$ and
$\mathbf{K},\mathbf{V}\in
\mathbb{R}^{T\times B\times N_m\times d}$.
Each occurrence of $\operatorname{Linear}$ and $\operatorname{BN}$ has
independent parameters. The affinity matrix
$\mathbf{R}_{a,n}=\mathbf{Q}_{a}\mathbf{K}^{\top}
\in\mathbb{Z}^{T\times B\times K\times N_m}$ associates each future
action position with the state-augmented memory.
$\operatorname{WTA}_{k}(\mathbf{R}_{a,n})$ converts these affinities
into a sparse binary routing matrix, allowing each action query to
retrieve a position-specific subset of visual, language, and robot-state
evidence. This retrieval is learned end-to-end through action
supervision. The routed features are temporally read out to produce the
continuous action chunk:
\begin{equation}
\begin{aligned}
\widehat{\mathcal{A}}_n =
\tanh\!\left(
\operatorname{Avg}_{T}\!\left[
\operatorname{Linear}_{a}\!\left(
\mathcal{SN}\!\left(
\operatorname{WTA}_{k}(\mathbf{R}_{a,n})\mathbf{V}
\right)\right)\right]\right)
\in
\mathbb{R}^{B\times K\times d_a}.
\end{aligned}
\label{eq:spkact}
\end{equation}
Here, $\operatorname{Linear}_{a}$ maps the routed spike features to the
action dimension $d_a$, while $\operatorname{Avg}_{T}$ performs
temporal readout over the spiking timesteps. The final $\tanh$ bounds
the continuous control commands. Consequently,
$\widehat{\mathcal{A}}_n =
[\widehat{\mathbf{a}}_n,\ldots,\widehat{\mathbf{a}}_{n+K-1}]$ predicts all $K$ future actions in
parallel.

\section{Experiments}
\subsection{Experiment Setup}

\textbf{Meta-World}: Meta-World~\citep{yu2020meta} comprises 50 diverse tabletop manipulation tasks performed by a Sawyer robot. We use the MT50 dataset released with SmolVLA~\citep{shukor2025smolvla}, containing 2,500 demonstrations with RGB observations, robot states, language instructions, and continuous actions. We jointly train a single policy on all tasks and evaluate it over 50 rollouts per task, reporting the average success rate based on task completion.

\textbf{LIBERO}: LIBERO~\citep{liu2023libero} consists of four suites: Spatial, Object, Goal, and Long. Each suite contains ten tasks specified by language instructions. We use the modified no-op RLDS datasets released with OpenVLA~\citep{kim2024openvla} and jointly train a single policy on all four suites. The policy predicts continuous 7-DoF actions in chunks of 12. We train the model for 80K steps, including 10K warm-up steps, with an effective batch size of 256. Following VLA-Adapter~\citep{wang2025vlaadapter}, we conduct 50 rollouts for each task and report success rates for each suite and across all 2,000 evaluation trials. To further assess zero-shot robustness under distribution shifts, we directly evaluate the same LIBERO-trained policy on LIBERO-Plus~\citep{fei2026liberoplus}, without any additional training or adaptation. LIBERO-Plus contains 10,030 perturbed task instances spanning seven dimensions: object layout, camera viewpoint, robot initial state, language instruction, lighting, background texture, and sensor noise.

\subsection{Main Results}

\begin{table*}[t]
    \centering
    \caption{
        Comparison on the Meta-World benchmark~\citep{yu2020meta}.
        Baseline success rates are taken from~\citep{shukor2025smolvla}.
        Our method is evaluated over 50 episodes per task.
    }
    \label{tab:metaworld_comparison}

    \small
    \setlength{\tabcolsep}{0.7pt}
    \renewcommand{\arraystretch}{1.15}

    \begin{tabular*}{\textwidth}{
        @{\extracolsep{\fill}}l*{9}{c}@{}
    }
        \toprule

        \multirow{3}{*}{\textbf{Method}}
        & \multirow{3}{*}{
            \shortstack{\textbf{Spike-}\\\textbf{driven}}
        }
        & \multicolumn{3}{c}{\textbf{Computation Efficiency}}
        & \multicolumn{5}{c}{\textbf{Meta-World Success Rate (\%)}} \\

        \cmidrule(lr){3-5}
        \cmidrule(lr){6-10}

        & & Params & FLOPs & Energy
        & \multirow{2}{*}{Easy}
        & \multirow{2}{*}{Medium}
        & \multirow{2}{*}{Hard}
        & \multirow{2}{*}{V. Hard}
        & \multirow{2}{*}{Avg.$\uparrow$} \\

        & & (B)$\downarrow$
          & (G)$\downarrow$
          & (mJ)$\downarrow$
        & & & & & \\

        \midrule
        TinyVLA~\citep{wen2024tinyvla}
        & $\times$ & 0.42 & 1259.1 & 2896.0
        & 77.6 & 21.5 & 11.4 & 15.8 & 31.6 \\

        $\pi_{0}$~\citep{black2024pi0}
        & $\times$ & 3.2 & 3069.8 & 7060.6
        & 71.8 & 48.2 & 41.7 & 30.0 & 47.9 \\

        SmolVLA (0.45B)~\citep{shukor2025smolvla}
        & $\times$ & 0.45 & 598.4 & 1376.4
        & 82.5 & 41.8 & 45.0 & 60.0 & 57.3 \\

        SmolVLA (2.25B)~\citep{shukor2025smolvla}
        & $\times$ & 2.25 & 4239.5 & 9751.1
        & 87.1 & 51.8 & 70.0 & 64.0 & 68.2 \\

        \midrule

        \rowcolor{oursblue}
        \textbf{Spike-driven VLA (Ours)}
        & $\checkmark$ & 0.15 & 10.5 & 11.6
        & 85.1 & 66.7 & 81.7
        & 56.0 & 72.4 \\
        \bottomrule
    \end{tabular*}
\end{table*}

As shown in Table~\ref{tab:metaworld_comparison}, Spike-driven VLA achieves an average success rate of 72.4\%, outperforming all compared ANN-based VLA models. It ranks first on Medium and Hard tasks, with success rates of 66.7\% and 81.7\%, respectively. Notably, it surpasses SmolVLA (2.25B) and SmolVLA (0.45B)~\citep{shukor2025smolvla} by 4.2 and 15.1\% while using $1/15$ and $1/3$ of their respective parameter counts. These results demonstrate that a compact spike-driven architecture can outperform larger ANN-based counterparts, with only 0.15B parameters, 10.5 GFLOPs, and an estimated inference energy of 11.6 mJ.

As shown in Table~\ref{tab:libero_comparison}, Spike-driven VLA achieves an average success rate of 92.4\% on LIBERO, with success rates above 95\% or equal to it on Spatial, Object, and Goal tasks and 82.4\% on Long tasks. It outperforms OpenVLA~\citep{kim2024openvla} and SmolVLA~\citep{shukor2025smolvla} by 15.9 and 3.6 percentage points, respectively, while closely matching DreamVLA (92.6\%)~\citep{zhang2025dreamvla}. Compared with $\pi_0$~\citep{black2024pi0}, it uses 95.3\% fewer parameters and reduces computational cost. We further evaluate robustness across seven perturbation dimensions on LIBERO-Plus~\citep{fei2026liberoplus}. As shown in Table~\ref{tab:libero_plus_comparison}, with only 0.15B parameters, Spike-driven VLA
achieves an overall success rate of 53.2\%, outperforming most ANN
baselines and remaining competitive with substantially larger VLA models. This performance indicates that Spike-driven VLA retains competitive robustness under distribution shifts. Overall, these results demonstrate Spike-driven VLA as a promising path toward resource-efficient embodied intelligence on edge platforms.

\begin{table*}[t]
    \centering
    \caption{
        Comparison on LIBERO~\citep{liu2023libero} benchmark. We report suite-level and overall average success rates based on 50 rollouts per task.
    }
    \label{tab:libero_comparison}
    \small
    \setlength{\tabcolsep}{3pt}
    \renewcommand{\arraystretch}{1.15}

    \begin{tabularx}{\textwidth}{
        @{}>{\raggedright\arraybackslash}X*{9}{c}@{}
    }
        \toprule

        \multirow{3}{*}{\textbf{Method}}
        & \multirow{3}{*}{
            \shortstack{\textbf{Spike-}\\\textbf{driven}}
        }
        & \multicolumn{3}{c}{Computational Efficiency}
        & \multicolumn{5}{c}{LIBERO Success Rate (\%)} \\

        \cmidrule(lr){3-5}
        \cmidrule(lr){6-10}

        & & Params & FLOPs & Energy
        & \multirow{2}{*}{Spa.}
        & \multirow{2}{*}{Obj.}
        & \multirow{2}{*}{Goal}
        & \multirow{2}{*}{Long}
        & \multirow{2}{*}{Avg.$\uparrow$} \\

        & & (B)$\downarrow$
          & (G)$\downarrow$
          & (mJ)$\downarrow$
        & & & & & \\

        \midrule

        OpenVLA~\citep{kim2024openvla}
        & $\times$ & 7.5 & 4158.3 & 9564
        & 84.7 & 88.4 & 79.2 & 53.7 & 76.5 \\

        $\pi_{0}$~\citep{black2024pi0}
        & $\times$ & 3.2 & 4371.9 & 10055
        & 96.8 & 98.8 & 95.8 & 85.2 & 94.2 \\

        $\pi_{0.5}$~\citep{physicalintelligence2025pi05}
        & $\times$ & 3.4 & 5014.2 & 11533
        & 98.8 & 98.2 & 98.0 & 92.4 & 96.9 \\

        CogVLA~\citep{li2025cogvla}
        & $\times$ & 8.3 & 3299.2 & 7588
        & 98.6 & 98.8 & 96.6 & 95.4 & 97.4 \\

        TurboVLA~\citep{xie2026turbovla}
        & $\times$ & 0.22 & 100.27 & 230.6
        & 99.2 & 99.8 & 97.4 & 94.2 & 97.7 \\

        SmolVLA~\citep{shukor2025smolvla}
        & $\times$ & 2.3 & 521.1 & 1199
        & 93.0 & 94.0 & 91.0 & 77.0 & 88.8 \\

        DreamVLA~\citep{zhang2025dreamvla}
        & $\times$ & 0.7 & 995.2 & 2289
        & 97.5 & 94.0 & 89.5 & 89.5 & 92.6 \\

        \midrule
        \rowcolor{oursblue}
        \textbf{Spike-driven VLA (Ours)}
        & $\checkmark$ & 0.15 & 15.68 & 20.9
        & 95.0 & 95.4 & 96.6 & 82.4 & 92.4 \\

        \bottomrule
    \end{tabularx}
     \vspace{-10pt}
\end{table*}

\begin{table*}[t]
    \centering
    \caption{
        Comparison on LIBERO-Plus~\citep{fei2026liberoplus} across seven perturbation dimensions.
    }
    \label{tab:libero_plus_comparison}

    \small
    \setlength{\tabcolsep}{0.5pt}
    \renewcommand{\arraystretch}{1.15}

    \begin{tabularx}{\textwidth}{
        @{}>{\raggedright\arraybackslash}X*{10}{c}@{}
    }
        \toprule

        \multirow{2}{*}{\textbf{Method}}
        & \multirow{2}{*}{
            \shortstack{\textbf{Spike-}\\\textbf{driven}}
        }
        & \multirow{2}{*}{
            \shortstack{\textbf{Params}\\(B)$\downarrow$}
        }
        & \multicolumn{8}{c}{
            \textbf{LIBERO-Plus Success Rate (\%)}
        } \\

        \cmidrule(lr){4-11}

        & & & Camera & Robot & Language & Light
        & Background & Noise & Layout & Avg.$\uparrow$ \\

        \midrule

        OpenVLA~\citep{kim2024openvla}
        & $\times$ & 7.5
        & 1.1 & 4.1 & 26.8 & 4.4
        & 25.3 & 19.3 & 31.6 & 16.1 \\

        NORA~\citep{hung2025nora}
        & $\times$ & 3.8
        & 4.0 & 41.1 & 67.0 & 31.0
        & 50.5 & 17.6 & 63.9 & 39.3 \\

        UniVLA~\citep{bu2025univla}
        & $\times$ & 7.0
        & 4.3 & 50.3 & 71.8 & 59.1
        & 80.0 & 25.3 & 34.3 & 46.4 \\

        $\pi_{0}$~\citep{black2024pi0}
        & $\times$ & 3.2
        & 15.8 & 6.6 & 61.0 & 79.6
        & 78.5 & 79.4 & 70.4 & 55.9 \\

        \midrule
        \rowcolor{oursblue}
        \textbf{Spike-driven VLA (Ours)}
        & $\checkmark$ & 0.15
        & 41.2 & 46.7 & 48.7 & 84.2
        & 61.3 &34.5 &  67.5 & 54.9 \\

        \bottomrule
    \end{tabularx}
\end{table*}


\subsection{Ablation Studies}
\begin{wraptable}{r}{0.35\columnwidth}
    \vspace{-20pt}
    \centering
    \caption{Ablation study of WTA.}
    \label{tab:wta_ablation}

    \small
    \setlength{\tabcolsep}{3pt}
    \renewcommand{\arraystretch}{1.08}

    \begin{tabularx}{\linewidth}{
        @{}>{\raggedright\arraybackslash}Xccc@{}
    }
        \toprule
        Suite
        & \shortstack{w/o WTA}
        & \shortstack{w/ WTA}
        & $\Delta$ \\
        \midrule

        Spatial & 90.2 & \textbf{95.0} & $+4.8$ \\
        Object  & \textbf{95.6} & 95.4 & $-0.2$ \\
        Goal    & 96.2 & \textbf{96.6} & $+0.4$ \\
        Long    & 78.6 & \textbf{82.4} & $+3.8$ \\

        \midrule
        Overall & 90.2 & \textbf{92.4} & $+2.2$ \\
        \bottomrule
    \end{tabularx}

    \vspace{-10pt}
\end{wraptable}
In this section, we conduct ablation studies on the perception and fusion stages of Spike-driven VLA. For perception, we compare spiking encoders with larger, full-precision ANN counterparts to assess the trade-off between task performance and computational efficiency. For fusion, we examine the effect of WTA on manipulation performance when visual tokens outnumber language tokens. The experimental settings and results are presented below.

\begin{table*}[t]
\centering
\caption{
Ablation of visual and language encoders on LIBERO. VE and LE denote the visual and language encoders, respectively. DINOv3-B~\citep{simeoni2025dinov3} and BERT~\citep{kenton2019bert} serve as the corresponding ANN baselines.
}
\label{tab:encoder_ablation}
\small
\setlength{\tabcolsep}{5pt}
\renewcommand{\arraystretch}{1.1}

\begin{tabular*}{\textwidth}{@{\extracolsep{\fill}}lll*{7}{r}@{}}
\toprule
\multirow{2}{*}{Variant}
& \multirow{2}{*}{VE}
& \multirow{2}{*}{LE}
& \multirow{2}{*}{\shortstack{Params\\(M)}}
& \multirow{2}{*}{\shortstack{Energy\\(mJ)}}
& \multicolumn{5}{c}{LIBERO Success Rate (\%)} \\
\cmidrule(lr){6-10}
& & & & & Spatial & Object & Goal & Long & Avg. \\
\midrule
B0 & DINOv3-B & BERT & 216.1 & 179.7
   & 98.6 & 99.2 & 93.2 & 94.0 & 96.3 \\
B1 & DINOv3-B & SIE & 216.1 & 173.0
   & 98.0 & 98.6 & 95.4 & 92.4 & 96.1 \\
B2 & SVE & BERT & 148.6 & 28.8
   & 95.8 & 98.2 & 95.6 & 80.8 & 92.6 \\
B3 & SVE & SIE & 148.6 & 20.9
   & 95.0 & 95.4 & 96.6 & 82.4 & 92.4 \\
\bottomrule
\end{tabular*}
\vspace{-10pt}
\end{table*}

\textbf{Perception}: As shown in Table~\ref{tab:encoder_ablation}, with the fusion module and action decoder fixed, replacing BERT with SIE reduces the average success rate by only 0.2\% (B0 vs. B1), whereas replacing DINOv3-B with SVE leads to a 3.7\% decrease (B1 vs. B3). This larger visual gap may reflect DINOv3-B's greater model capacity and richer visual priors acquired through large-scale pretraining. These results suggest that visual perception accounts for most of the performance gap, while SIE achieves performance comparable to BERT and retains sufficient capacity to distinguish task instructions on LIBERO.

\begin{figure*}[!htpb]
  \centering
  \includegraphics[width=\linewidth]{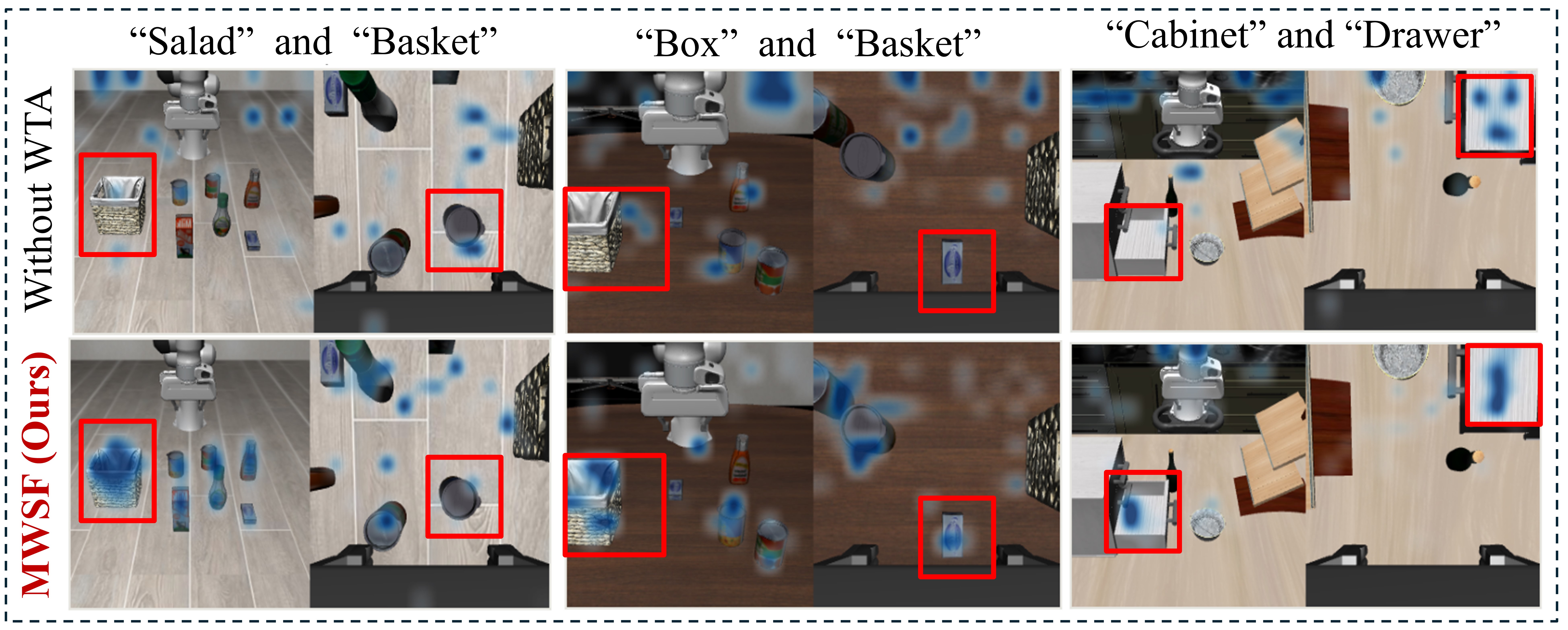}
\caption{
 Attention heatmaps showing that our MWSF module can better focus on instruction-relevant visual regions, highlighted by red boxes. 
}
  \label{fig:4}
\vspace{-5pt}
\end{figure*}

\textbf{Fusion}: As shown in Table~\ref{tab:wta_ablation},
WTA increases the overall success rate from 90.2\% to 92.4\%, with gains of 4.8\% and 3.8\% on the Spatial and Long suites, respectively. The heatmaps in Fig.~\ref{fig:4} further show that MWSF with WTA focuses attention on instruction-relevant visual regions (red boxes) while suppressing background responses. These findings suggest that WTA strengthens instruction-guided visual grounding, contributing to improved performance in long-horizon manipulation.

\section{Conclusion}
In this work, we present Spike-driven VLA, the first spiking VLA framework to support end-to-end direct training for robotic manipulation. By integrating SVE, SIE, the MWSF module, and SpikeACT, it achieves competitive performance on LIBERO and Meta-World with fewer parameters and operations than representative ANN-based methods. These results demonstrate that directly trained spiking architectures offer a viable approach to resource-efficient embodied intelligence. Future work will extend the framework to broader tasks and diverse robotic platforms to advance the development of neuromorphic embodied agents.


\bibliography{iclr2027_conference}
\bibliographystyle{iclr2027_conference}

\appendix

\section{Spiking Visual Encoder}
\label{app:sve}
\paragraph{Network architecture.}
Our spiking visual encoder adopts a three-stage pyramidal architecture. The first two stages contain one and two convolutional mixer blocks, respectively. These blocks extract local features through spatial and channel mixing. The third stage comprises seven spiking self-attention blocks to model global spatial dependencies. Each attention block combines spike-based feature transformations with residual connections. Spike Patch Embedding layers progressively reduce spatial resolution and increase channel width. For an input of size $H \times W$, the three stages produce feature maps of size $C \times H/4 \times W/4$, $2C \times H/8 \times W/8$, and $4C \times H/16 \times W/16$, respectively. Here, $C$ denotes the base channel width. This hierarchy extracts fine-grained local features at high resolutions and captures global spatial dependencies with fewer visual tokens.

\begin{wraptable}{r}{0.48\textwidth}
    \centering
    \vspace{-20pt}
    \caption{ImageNet-1K pretraining settings for the spiking visual encoder.}
    \label{tab:sve_pretraining}
    \small
    \setlength{\tabcolsep}{4pt}
    \renewcommand{\arraystretch}{1.1}
    \resizebox{\linewidth}{!}{%
        \begin{tabular}{@{}ll@{}}
            \toprule
            Hyperparameter & Setting \\
            \midrule
            Time steps & 4 \\
            Training epochs & 200 \\
            Input resolution & $224 \times 224$ \\
            Batch size & 600 \\
            Optimizer & LAMB \\
            Base learning rate & $6 \times 10^{-4}$ \\
            Warmup epochs & 10 \\
            Layer-wise LR decay & 1.0 \\
            Weight decay & 0.05 \\
            Mixup & 0 \\
            CutMix & 0 \\
            Label smoothing & 0.1 \\
            Hardware & $4 \times$ NVIDIA A800 \\
            \bottomrule
        \end{tabular}%
    }
     \vspace{-10pt}
\end{wraptable}
\paragraph{ImageNet-1K pretraining.}
To improve visual generalization, we pretrain SVE on ImageNet-1K before training Spike-driven VLA for robotic grasping. The dataset contains approximately 1.3 million training images and 50,000 validation images across 1,000 classes. A classification head maps visual features to 1,000 class logits. Training uses cross-entropy loss with a label-smoothing coefficient of $0.1$. Surrogate gradients enable backpropagation through spiking activations.

Pretraining is implemented in PyTorch with distributed data parallelism on four NVIDIA A800 GPUs. We train for 200 epochs with $224 \times 224$ inputs and a batch size of 600. The LAMB optimizer uses a base learning rate of $6 \times 10^{-4}$ and a weight decay of $0.05$. The first 10 epochs are used for learning-rate warmup. The layer-wise learning-rate decay factor is $1.0$, assigning the same multiplier to all backbone layers. Mixup and CutMix are disabled. Table~\ref{tab:sve_pretraining} summarizes the pretraining configuration.

\paragraph{Dual-View Encoding and Joint Fine-Tuning.}
After pretraining, we remove the classification head and retain the visual backbone. Main-view and wrist-view observations are processed by the same backbone with shared weights. For each view, the final $14 \times 14$ feature map is flattened into 196 spatial tokens. These tokens are linearly projected to the common embedding dimension $d=256$ and normalized. We then concatenate the tokens from both views along the token dimension, yielding $\mathbf{Z}_{v,n}\in\mathbb{R}^{T \times B \times 392 \times 256}$ for the LIBERO task. The backbone is jointly fine-tuned with the cross-modal fusion and action-decoding modules for language-conditioned robotic manipulation.

\section{Spike Instruction Encoder}
\label{app:sie}
The SIE $E_{\ell}$ encodes a language instruction $\mathcal{L}$ into task-conditioned spiking tokens for visual perception. It first tokenizes $\mathcal{L}$ into $\operatorname{Tok}(\mathcal{L}) \in \mathbb{Z}^{B \times N_{\ell}}$. A spiking BERT backbone~\citep{liangspikinglm,zhousmoothspike} then extracts contextual representations $\mathbf{Z}^{\mathrm{text}}$, which are projected from $d_{\ell}$ to the shared VLA embedding dimension $d$ by a dimensional adapter:
\begin{equation}
\mathbf{Z}_{\ell}
= E_{\ell}\!\left(\operatorname{Tok}(\mathcal{L})\right)
= \operatorname{BN}\!\left(
    \operatorname{Linear}_{\ell}\!\left(\mathbf{Z}^{\mathrm{text}}\right)
  \right),
\qquad
\mathbf{Z}_{\ell}
\in \mathbb{R}^{T \times B \times N_{\ell} \times d}.
\label{eq:sie}
\end{equation}

We instantiate the backbone with SmoothSpike~\citep{zhousmoothspike}, which comprises 12 encoder layers with 12 attention heads per layer. It uses a hidden dimension of $d_{\ell}=768$, an MLP dimension of $3{,}072$, and $T=4$ simulation steps. The backbone is pretrained using masked language modeling~\citep{kenton2019bert} and remains frozen during Spike-driven VLA training. Consequently, the manipulation loss updates only $\operatorname{Linear}_{\ell}$ and the subsequent cross-modal fusion and action-decoding modules. In the following, we detail the pretraining data, the SmoothSpike method, and the SIE architecture.

\paragraph{Pre-training corpus.}
SmoothSpike follows the BERT masked-language-modeling (MLM) paradigm~\citep{kenton2019bert}.
The pre-training corpus is assembled from publicly available English text collections accessed through the Hugging Face Datasets library: TinyStories,\footnote{\url{https://huggingface.co/datasets/roneneldan/TinyStories}} BookCorpus,\footnote{\url{https://huggingface.co/datasets/bookcorpus}} CC-News,\footnote{\url{https://huggingface.co/datasets/vblagoje/cc_news}} OpenWebText,\footnote{\url{https://huggingface.co/datasets/Skylion007/openwebtext}} and English Wikipedia.\footnote{\url{https://huggingface.co/datasets/wikimedia/wikipedia}}
Each source is stored as a Hugging Face \texttt{Dataset} (or \texttt{DatasetDict}) with a free-text field.
When a corpus provides a \texttt{text} column, we use it directly; otherwise, we take the first available text-bearing column.
For sources without an official validation split, the first $\min(1000,N_{\mathrm{train}})$ raw documents of that source are reserved as its validation contribution.
Train splits from all five sources are concatenated into a single training pool, and the corresponding validation contributions form a held-out validation pool. No additional document-level filtering, deduplication, or language identification is applied beyond what is already present in the upstream releases.

We tokenize the corpus using the bert-base-uncased WordPiece tokenizer with a vocabulary of $30{,}522$ tokens. Within each preprocessing batch, tokenized documents are concatenated and divided into non-overlapping blocks of $L=128$ tokens. Incomplete trailing blocks are discarded. Each block stores token IDs, token-type IDs, attention masks, and special-token masks. The processed corpus contains $137{,}271{,}688$ training blocks and $1{,}627{,}657$ validation blocks, corresponding to approximately $17.6$ billion training tokens. MLM targets are generated dynamically by Hugging Face's DataCollatorForLanguageModeling with a masking probability of $15\%$. Special tokens are excluded from masking, and new masking patterns are sampled in each epoch. The pretrained backbone is frozen to provide semantic features $\mathbf{Z}^{\mathrm{text}}$ for SIE during VLA training and inference. Instructions from LIBERO and Meta-World use the same tokenizer and are encoded as variable-length sequences of $N_{\ell}\le128$ tokens.

\begin{table}[h]
\centering
\caption{Training hyperparameters of the SmoothSpike backbone used as SIE.
The VLA column summarizes its configuration in Spike-driven VLA,
where the backbone remains frozen.}
\label{tab:sie_hparams}
\small
\setlength{\tabcolsep}{4pt}
\renewcommand{\arraystretch}{1.12}
\begin{tabular*}{\linewidth}{@{\extracolsep{\fill}}lcc@{}}
\toprule
Hyperparameter & MLM pre-training & Spike-driven VLA training \\
\midrule
Objective & MLM (15\% masking) & Frozen encoder \\
Tokenizer & \texttt{bert-base-uncased} & Same \\
Maximum sequence length & $128$ & $N_{\ell} \leq 128$ \\
Optimizer & AdamW & -- \\
Peak learning rate & $2 \times 10^{-4}$ & -- \\
Weight decay & $0$ & -- \\
LR schedule & Linear, 5k warm-up & -- \\
Per-device / global batch size & $64$ / $512$ & -- \\
Time steps $T$ & $4$ & $4$ \\
Tokens seen & $36.0$B & -- \\
Trainable modules & Full SmoothSpike
                  & $\operatorname{Linear}_{\ell}$, BN only \\
\bottomrule
\end{tabular*}
\end{table}

\begin{table*}[t]
\centering
\caption{Configuration of the Spike Instruction Encoder (SIE) used in Spike-driven VLA.
The backbone is SmoothSpike-12-768~\citep{zhousmoothspike}.
$N_{\ell}$ is the instruction length ($N_{\ell}\le 128$).
$\mathbf{H}_{1}$ is shared across residual branches; $\mathbf{H}_{2}$ is per-layer on the value branch; $\mathbf{H}_{3}$ is a $4$-block diagonal transform in the MLP.
Query/key LIF neurons are not transformed.}
\label{tab:sie_arch}
\small
\setlength{\tabcolsep}{5pt}
\renewcommand{\arraystretch}{1.15}
\resizebox{\textwidth}{!}{%
\begin{tabular}{c|c|ccc|c}
\hline
stage & \# Tokens & \multicolumn{3}{c|}{Layer Specification} & SIE \\
\hline
\multirow{5}{*}{Emb.}
  & \multirow{5}{*}{$N_{\ell}$}
  & \multicolumn{2}{c}{Token / position / segment Emb.}
  & Dim
  & $768$ \\
\cline{3-6}
  & & \multicolumn{2}{c}{Vocabulary / max position}
  & --
  & $30522$ / $512$ \\
\cline{3-6}
  & & \multicolumn{2}{c}{Embedding Norm}
  & --
  & LayerNorm \\
\cline{3-6}
  & & \multicolumn{2}{c}{Shared orthogonal encoding}
  & --
  & $\mathbf{H}_{1}$ \\
\cline{3-6}
  & & \multicolumn{2}{c}{Temporal repeat}
  & $T$
  & $4$ \\
\hline
\multirow{12}{*}{\shortstack{Encoder\\block}}
  & \multirow{12}{*}{$N_{\ell}$}
  & \multirow{6}{*}{SSA}
  & RMSNorm + LIF + $\mathbf{H}_{1}^{\top}$
  & $\tau$
  & $2.0$ \\
\cline{4-6}
  & & & $Q/K$ Linear + RMSNorm + LIF
  & Heads
  & $12$ \\
\cline{4-6}
  & & & $V$ Linear + $\mathbf{H}_{2}$ + RMSNorm + LIF
  & Dim
  & $768$ \\
\cline{4-6}
  & & & Spike-driven attn.\ $(QK^{\top}\odot M)sV$
  & $s$
  & $1/8$ \\
\cline{4-6}
  & & & LIF + $\mathbf{H}_{2}^{\top}$ + $W_{o}$ + $\mathbf{H}_{1}$
  & $v_{\mathrm{th}}$
  & $0.5$ \\
\cline{4-6}
  & & & Residual + dropout
  & $p$
  & $0.1$ \\
\cline{3-6}
  & & \multirow{5}{*}{MLP}
  & RMSNorm + LIF + $\mathbf{H}_{1}^{\top}$
  & $\tau$
  & $2.0$ \\
\cline{4-6}
  & & & Up-projection
  & Dim
  & $768\to 3072$ \\
\cline{4-6}
  & & & Block-diagonal $\mathbf{H}_{3}$
  & \# blocks
  & $4$ \\
\cline{4-6}
  & & & LIF + $\mathbf{H}_{3}^{\top}$ + down-projection + $\mathbf{H}_{1}$
  & Dim
  & $3072\to 768$ \\
\cline{4-6}
  & & & Residual + dropout
  & $p$
  & $0.1$ \\
\cline{3-6}
  & & \multicolumn{2}{c}{\# Blocks}
  & --
  & $12$ \\
\hline
\multirow{3}{*}{Head}
  & \multirow{3}{*}{$N_{\ell}$}
  & \multicolumn{2}{c}{MLM readout (pre-training)}
  & --
  & $\mathrm{mean}_{T}$ + $\mathbf{H}_{1}^{\top}$ \\
\cline{3-6}
  & & \multicolumn{2}{c}{SIE features $\mathbf{Z}^{\mathrm{text}}$}
  & Dim
  & $T\times N_{\ell}\times 768$ \\
\cline{3-6}
  & & \multicolumn{2}{c}{VLA adapter $\operatorname{BN}(\operatorname{Linear}_{\ell}(\cdot))$}
  & Dim
  & $768\to d$ \\
\hline
\end{tabular}%
}
\end{table*}

\paragraph{Training protocol.}
SmoothSpike is trained directly as an SNN, without ANN-to-SNN conversion.
Each forward pass resets neuronal states before the sequence is consumed.
LIF neurons run in multi-step mode with membrane time constant $\tau=2.0$ and detached reset; the default firing threshold is $1.0$, except for the attention-output LIF whose threshold is $0.5$.
Learnable orthogonal transforms $\widetilde{\mathbf{H}}$ are initialized as randomized Hadamard matrices and projected onto the orthogonal group at every forward pass by a quintic Newton--Schulz iteration with $K=5$ steps and coefficients $(a,b,c)=(3.4445,-4.7750,2.0315)$.
$\mathbf{H}_{1}$ is shared across residual-connected branches, $\mathbf{H}_{2}$ is applied to the value branch, and $\mathbf{H}_{3}$ is a block-diagonal MLP transform with four $768\times 768$ blocks.
Query/key LIF neurons are left untransformed.
After language-model training, all transforms are fused into adjacent linear weights, so SIE inference introduces no extra matmuls and remains spike-driven.

MLM pre-training minimizes the cross-entropy on masked tokens with AdamW (peak learning rate $2\times 10^{-4}$, $\beta_{1}=0.9$, $\beta_{2}=0.999$, no weight decay).
The learning rate warms up linearly for $5{,}000$ steps and then decays linearly to zero.
Training uses $8$ GPUs, $64$ sequences per device, no gradient accumulation (global batch size $512$), and maximum length $128$.
The run consumes approximately $36.0$ billion tokens.
Hidden-state and attention dropout are both $0.1$, and training is performed in full precision. Table~\ref{tab:sie_hparams} summarizes the language-backbone hyperparameters.

\paragraph{Architecture.}

Table~\ref{tab:sie_arch} specifies the SIE architecture. Although the table follows the stage-wise convention used for hierarchical spike-driven Transformers, SIE preserves the instruction token count $N_\ell$ throughout.
It adopts a pre-norm spiking BERT encoder, with LIF neurons implementing the spiking operation $\mathcal{SN}(\cdot)$. Token, position, and segment embeddings are summed and normalized, right-multiplied by the shared orthogonal transform $\mathbf H_1$, and presented over $T$ timesteps. Each encoder block consists of SSA and a spiking MLP.In SSA, RMSNorm and a projection LIF generate spikes that are mapped back by $\mathbf H_1^\top$ before the $\mathbf Q$, $\mathbf K$, and $\mathbf V$ projections. The $\mathbf Q$ and $\mathbf K$ branches each apply RMSNorm and $\mathcal{SN}(\cdot)$, whereas the $\mathbf V$ branch applies $\mathbf H_2$ before RMSNorm and $\mathcal{SN}(\cdot)$. Following the SSA notation introduced above, attention is computed without softmax as:
\begin{equation}
\mathbf{Score}
= s \cdot \bigl((\mathbf Q \times \mathbf K^\top) \odot \mathbf M\bigr),
\quad
\mathbf{Attn}
= \mathcal{SN}(\mathbf{Score} \times \mathbf V),
\end{equation}
where $\mathbf M$ denotes the attention mask and $s=1/\sqrt{d_h}=1/8$. The spiking attention output $\mathbf{Attn}$ then passes through $\mathbf H_2^\top$, the output projection $\mathbf W_o$, and
$\mathbf H_1$ before the residual addition. The MLP applies RMSNorm and $\mathcal{SN}(\cdot)$, followed by
$\mathbf H_1^\top$, an up-projection from $768$ to $3072$ dimensions, and the block-diagonal transform $\mathbf H_3$, which comprises four $768\times768$ blocks. A second $\mathcal{SN}(\cdot)$ operation is followed by $\mathbf H_3^\top$, a down-projection to $768$ dimensions, and $\mathbf H_1$ before the residual addition.
For MLM pre-training, the encoder outputs are averaged over $T$, mapped back by $\mathbf H_1^\top$, and passed to the MLM head. In Spike-driven VLA, the MLM-pretrained SmoothSpike encoder is frozen, and each instruction is processed over $T=4$ timesteps to match the visual stream. We retain the timestep-resolved encoder features
$\mathbf Z^{\mathrm{text}}\in
\mathbb R^{T\times B\times N_\ell\times d_\ell}$
without averaging over time, where $B$ denotes the batch size and$d_\ell=768$ is the encoder feature dimension.
A linear projection followed by batch normalization maps each token feature to the shared feature dimension $d=256$:
\begin{equation}
\mathbf Z_\ell
= \text{BN}\bigl(\mathrm{Linear}(\mathbf Z^{\mathrm{text}})\bigr)
\in \mathbb R^{T\times B\times N_\ell\times d}.
\end{equation}
These features preserve the distinctions among task instructions and provide the semantic cues used by MWSF module to select task-relevant visual information.

\begin{figure*}[!htpb]
  \centering
  \includegraphics[width=\linewidth]{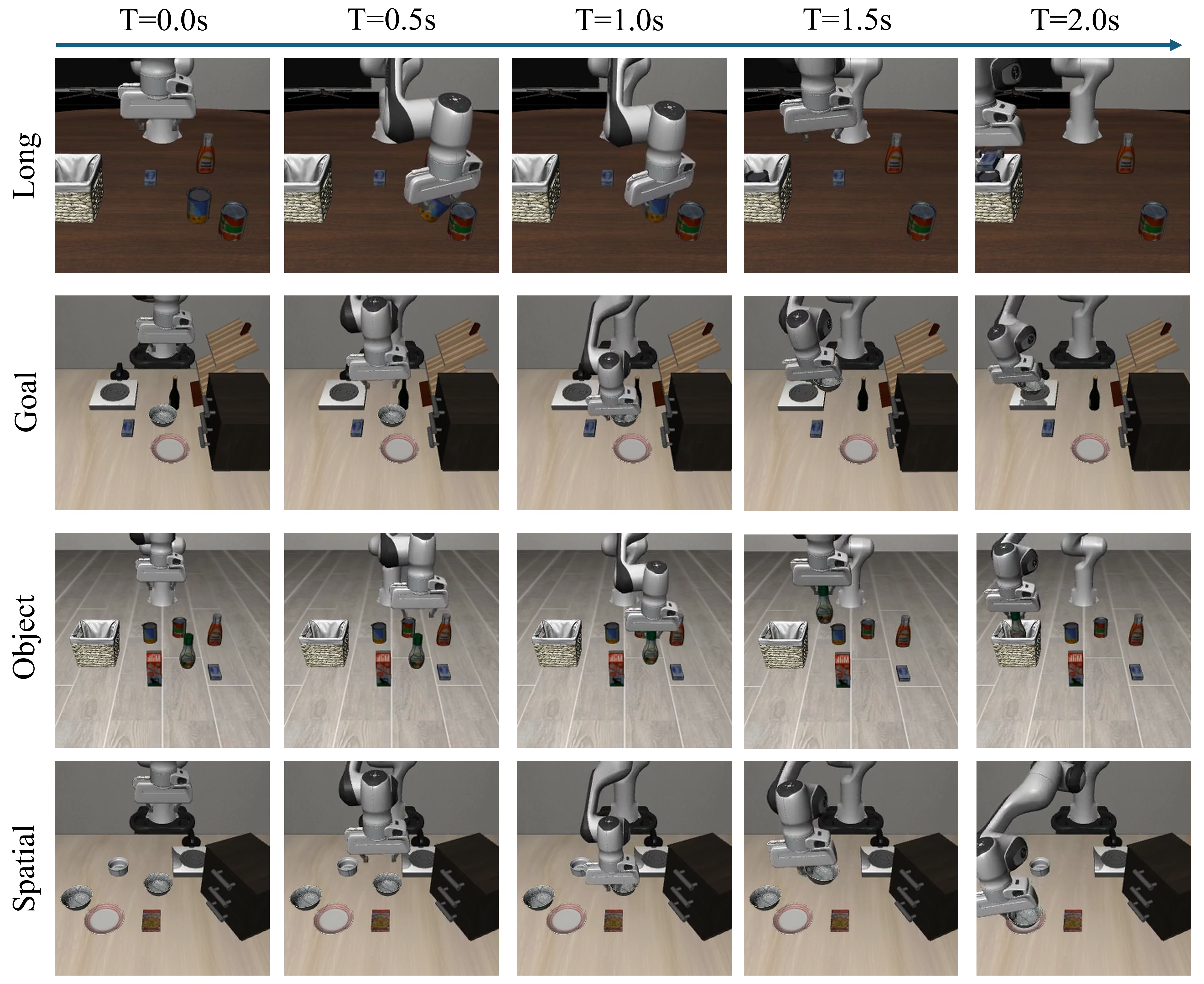}
  \caption{
    Successful Spike-driven VLA rollouts on LIBERO.
    Rows (a)--(d) show four tasks. Columns show frames sampled at
    T= $0$, $0.5$, $1.0$, $1.5$, and $2.0$.
  }
  \label{fig:app_libero_keyframes}
\end{figure*}

\section {Training Details of Spike-driven VLA}
\label{app:training_pipeline}
\paragraph{Training flow.}
For each benchmark, Spike-driven VLA is trained as a single policy across its tasks. The SmoothSpike BERT backbone remains frozen, while SVE, the visual and text projections, the MWSF layers, the state encoder, and the SpikeACT decoder are optimized jointly. Before policy training, the text projection is calibrated on benchmark-specific instructions to initialize its weights. For each sample, SVE encodes the RGB observations into $N_v$ visual tokens, while SmoothSpike BERT encodes the instruction into $N_{\ell}$ text tokens. Both streams are projected to the shared feature dimension $d$. The MWSF layers fuse them into $N_v+N_{\ell}$ tokens. Concatenating $N_s$ encoded state tokens forms a memory of shape $T\times B\times N_m\times d$, where $N_m=N_v+N_{\ell}+N_s$. SpikeACT uses this memory to predict an action chunk of shape $B\times K\times d_a$, where $K$ is the action horizon and $d_a$ is the action dimension. The neuronal simulation length $T$ is distinct from the action horizon $K$.

For both benchmarks, behavior cloning uses a masked L1 loss between
predicted and demonstrated action chunks. The loss is averaged over
valid action elements, with equal weight assigned to each action
dimension. Text-projection calibration is performed before policy
training and serves only as initialization. We use neither staged
objective switching nor ANN-to-SNN conversion.

On LIBERO, $K=12$ and $d_a=7$, giving action chunks of shape
$B\times12\times7$. The text projection is calibrated on 40 LIBERO
instructions. Table~\ref{tab:spike_vla_training} reports the
optimization settings for the LIBERO policy. On Meta-World MT50, the policy predicts 12-step chunks in the
benchmark's four-dimensional action space, giving a shape of
$B\times12\times4$. Its text projection is calibrated using MT50
instructions.

\begin{table}[t]
\centering
\caption{Training configuration of Spike-driven VLA on LIBERO.}
\label{tab:spike_vla_training}
\small
\setlength{\tabcolsep}{5pt}
\renewcommand{\arraystretch}{1.12}
\begin{tabularx}{\linewidth}{
    @{}l >{\raggedright\arraybackslash}X@{}
}
\toprule
Item & Configuration \\
\midrule
Effective batch size
    & $8$ GPUs $\times$ $32$ samples/GPU
      $\times$ $1$ accumulation step $=256$ \\
Training budget
    & $80{,}000$ updates; $20{,}480{,}000$ sample exposures \\
Optimizer
    & AdamW, $\beta=(0.9,0.95)$, $\epsilon=10^{-8}$ \\
Visual backbone LR
    & $1\times10^{-5}$ \\
Other trainable parameters LR
    & $5\times10^{-5}$ \\
LR schedule
    & $10{,}000$ steps of linear warm-up,
      followed by a constant LR \\
Gradient clipping
    & Global gradient norm capped at $1$ \\
EMA
    & Decay $0.999$; updated after every optimizer step \\
Policy precision
    & FP32; highest float32 matrix-multiplication precision;
      TF32 disabled \\
\midrule
$\mathcal{L}_{\mathrm{BC}}$
    & Masked L1 loss on $B\times12\times7$ action chunks;\\
\bottomrule
\end{tabularx}
\end{table}

\section{Theoretical Energy Estimation}
\label{app:energy_estimation}
We estimate arithmetic energy per policy forward pass, which produces a 12-step action chunk. The estimate includes SmoothSpike, whose weights are frozen during policy training but whose computation remains part of inference. Direct encoding supplies continuous-valued images to the first convolution, so its operations are counted as MACs. In subsequent spike-driven layers, synaptic accumulations occur only in response to nonzero input spikes.

For a spike-driven convolutional or linear layer $\ell$, let $M_\ell$ denote the dense MAC count per execution, $r_{\ell,t}$ the input firing rate at its $t$-th execution, and $T_\ell$ the number of executions in one forward pass. We estimate its synaptic operations as:
\begin{equation}
\widehat{\mathrm{SOP}}_\ell
\simeq M_\ell\sum_{t=1}^{T_\ell}r_{\ell,t}.
\label{eq:layer_sops}
\end{equation}
Counting actual executions avoids applying the four simulation steps again to operations performed after temporal aggregation. Spike-triggered accumulations in attention are counted as ACs, while operations on continuous-valued inputs are counted as MACs. Let $N_{\mathrm{AC}}$ and $N_{\mathrm{MAC}}$ be the respective totals over the complete forward pass. The theoretical computation energy is:
\begin{equation}
\widehat E_{\mathrm{SpikeVLA}}
=e_{\mathrm{AC}}N_{\mathrm{AC}}
+e_{\mathrm{MAC}}N_{\mathrm{MAC}}.
\label{eq:spike_vla_energy}
\end{equation}
Following the 45-nm reference model for 32-bit arithmetic~\citep{zhou2023spikformer,zhou2024qkformer}, we use
$e_{\mathrm{AC}}=0.9\,\mathrm{pJ}$ and $e_{\mathrm{MAC}}=4.6\,\mathrm{pJ}$. Where applicable, batch normalization and SmoothSpike's orthogonal transforms are folded into adjacent weights before counting operations. The estimate excludes
memory access and non-synaptic overhead.

\section{Evaluation on LIBERO and Meta-World}
\label{app:bmatrix_results}
\paragraph{LIBERO.}
We evaluate a single policy on the Spatial, Object, Goal, and Long
suites. Each demonstration provides workspace and wrist RGB observations,
an instruction, robot states, and actions. The two camera views are
processed by the shared SVE to produce 392 visual tokens, while
SmoothSpike encodes the instruction into 21 text tokens. At each
trajectory step, we construct a 12-step relative-action target and mask
padded actions near the trajectory end. Each suite contains ten tasks,
with 50 evaluation trials per task, giving 2,000 trials overall.
Evaluation uses EMA weights, FP32 inference, and seed 7; each predicted
action chunk is executed open loop for 12 steps.
Figure~\ref{fig:app_libero_keyframes} shows four successful rollouts
generated by the policy.

\begin{table*}[t]
    \centering
    \caption{
        Success rates (\%) on LIBERO-Plus. Parentheses give successful
        and total rollout counts. Overall rates are computed from pooled
        rollout counts.
    }
    \label{tab:libero_plus_breakdown}
    \small
    \setlength{\tabcolsep}{3pt}
    \renewcommand{\arraystretch}{1.3}
    \begin{tabular*}{\textwidth}{
        @{\extracolsep{\fill}}l*{8}{c}@{}
    }
        \toprule
        \textbf{Suite}
        & \textbf{Overall}
        & \shortstack{Background\\Textures}
        & \shortstack{Robot Initial\\States}
        & \shortstack{Camera\\Viewpoints}
        & \shortstack{Language\\Instructions}
        & \shortstack{Sensor\\Noise}
        & \shortstack{Object\\Layout}
        & \shortstack{Light\\Conditions} \\
        \midrule

        Spatial & 65.7
        & \shortstack{72.1\\(186/258)}
        & \shortstack{57.7\\(202/350)}
        & \shortstack{63.0\\(237/376)}
        & \shortstack{40.0\\(156/390)}
        & \shortstack{48.4\\(170/351)}
        & \shortstack{87.0\\(335/385)}
        & \shortstack{99.7\\(291/292)} \\[3pt]

        Object & 56.0
        & \shortstack{70.6\\(175/248)}
        & \shortstack{42.2\\(168/398)}
        & \shortstack{42.9\\(170/396)}
        & \shortstack{58.2\\(206/354)}
        & \shortstack{32.7\\(138/422)}
        & \shortstack{65.8\\(265/403)}
        & \shortstack{97.3\\(289/297)} \\[3pt]

        Goal & 54.0
        & \shortstack{69.4\\(195/281)}
        & \shortstack{50.9\\(208/409)}
        & \shortstack{46.8\\(191/408)}
        & \shortstack{35.4\\(145/410)}
        & \shortstack{43.5\\(165/379)}
        & \shortstack{60.9\\(259/425)}
        & \shortstack{84.6\\(236/279)} \\[3pt]

        Long & 37.6
        & \shortstack{36.0\\(104/289)}
        & \shortstack{37.2\\(146/393)}
        & \shortstack{14.6\\(61/419)}
        & \shortstack{63.2\\(242/383)}
        & \shortstack{17.6\\(79/449)}
        & \shortstack{54.5\\(170/312)}
        & \shortstack{52.9\\(145/274)} \\
        \bottomrule
    \end{tabular*}
\end{table*}
Across seven LIBERO-Plus perturbation categories, Spike-driven VLA
achieves 53.2\% success over 10,030 rollouts
(Table~\ref{tab:libero_plus_breakdown}). Success is highest under
lighting (84.2\%) and object-layout changes (67.5\%), but lowest under
sensor noise (34.5\%) and camera-viewpoint changes (41.2\%). The Long
suite is particularly challenging (37.6\% overall), dropping to 14.6\%
under viewpoint changes and 17.6\% under sensor noise.

\begin{figure*}[!htpb]
  \centering
  \includegraphics[width=\linewidth]{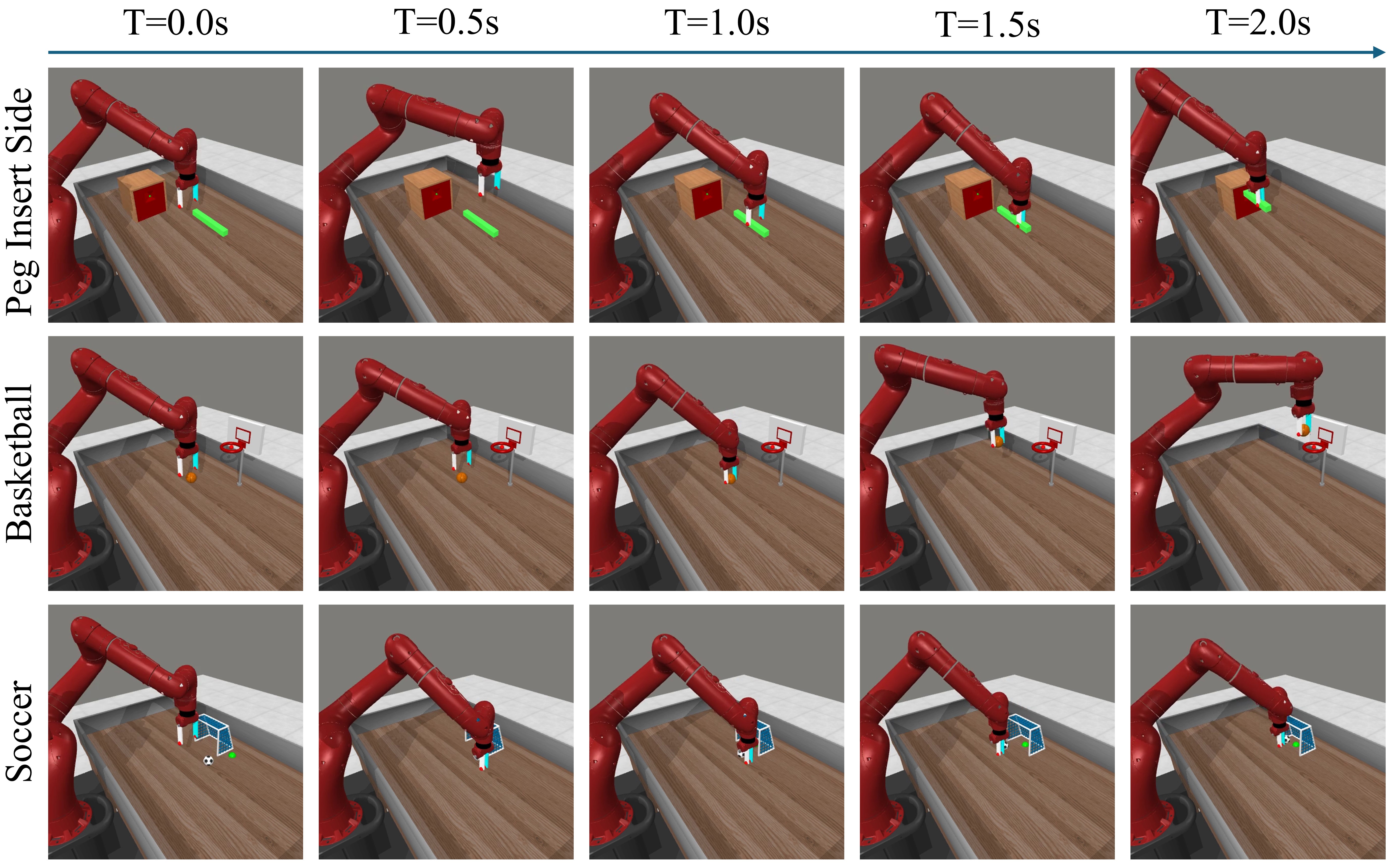}
  \caption{
    Successful Spike-driven VLA rollouts on Meta-World MT50.
    Rows (a)--(d) show four tasks. Columns show frames sampled at T= $0$, $0.5$, $1.0$, $1.5$, and $2.0$.
  }
  \label{fig:app_metaworld_keyframes}
\end{figure*}

\paragraph{Meta-World MT50.}
For MT50, we train a single policy across all 50 tasks using
50 expert demonstrations per task, yielding 2,500 trajectories
in total. Rendered observations, task descriptions, robot states,
and expert actions are aligned at each simulation step.
The action sequences are divided into 12-step targets, with
padded positions excluded from the training loss.
We evaluate the policy over 50 trials per task, yielding
2,500 evaluation trials in total.
Figure~\ref{fig:app_metaworld_keyframes} presents successful
rollouts from representative MT50 tasks.

\end{document}